\documentclass[letterpaper, 10 pt, conference]{ieeeconf}  

\IEEEoverridecommandlockouts                              

\usepackage{amsmath} 
\usepackage[pdftex]{graphicx}
\usepackage[Option]{overpic}
\usepackage{booktabs}
\usepackage[caption=false]{subfig}
\usepackage{multirow}
\usepackage{adjustbox}
\usepackage{float}
\usepackage{bbding}
\usepackage{amssymb}
\usepackage{pifont}
\usepackage[Algorithm]{algorithm}
\usepackage{algorithmic}
\newcommand{\cmark}{\ding{51}}%
\newcommand{\xmark}{\ding{55}}%

\title{\LARGE \bf PDC: Piecewise Depth Completion utilizing Superpixels}

\author{Dennis Teutscher$^{1}$, Patrick Mangat$^{1}$ and Oliver Wasenm\"uller$^{1}$
\thanks{$^{1}$Mannheim University of Applied Sciences, Germany}
\thanks{{\tt\small dennis.teutscher@stud.hs-mannheim.de}}%
\thanks{{\tt\small p.mangat@hs-mannheim.de}}%
\thanks{{\tt\small o.wasenmueller@hs-mannheim.de}}%
}

\begin{document}

\maketitle
\thispagestyle{empty}
\pagestyle{empty}

\begin{abstract}

Depth completion from sparse LiDAR and high-resolution RGB data is one of the foundations for autonomous driving techniques. 
Current approaches often rely on CNN-based methods with several known drawbacks: flying pixel at depth discontinuities, overfitting to both a given data set as well as error metric, and many more.
Thus, we propose our novel Piecewise Depth Completion (PDC), which works completely without deep learning.
PDC segments the RGB image into superpixels corresponding the regions with similar depth value.
Superpixels corresponding to same objects are gathered using a cost map.
At the end, we receive detailed depth images with state of the art accuracy.
In our evaluation, we can show both the influence of the individual proposed processing steps and the overall performance of our method on the challenging KITTI dataset.

\end{abstract}


\section{Introduction}

Currently, the transportation industry attaches great importance to the development of autonomous driving.
Taking into account 3D Light Detection And Ranging (LiDAR) sensors becomes increasingly important. 
This sensor allows to determine sparse distance measurements at high accuracy. 
For further higher-level processing this sparse information needs to be transformed into a dense depth image. 
The method of creating a dense depth map from a sparse LiDAR input is called depth completion.

Recent approaches use different architectures of Convolutional Neural Networks (CNN) together with a high-resolution RGB image of the same scene for depth completion.
Even though these methods are able to produce reasonable results, there are still some issues. 
The inference of depth values using convolutions creates so called flying pixels (see Figure \ref{teaser} (e)) -- especially in regions with discontinuities.
Flying pixels are incorrect 3D points that connect a surface in the foreground to a surface in the background, even though there should be no geometry.
Sometimes they are also called ghost points or slopes \cite{wasen}.
These flying pixels corrupt the depth image and yield an erroneous representation of reality. 
Additionally, CNN-based methods tend to be optimized for a particular data set and error metric, so generalization is sometimes not given.

\begin{figure*}[t]
    \subfloat{\includegraphics[width =0.38\textwidth]{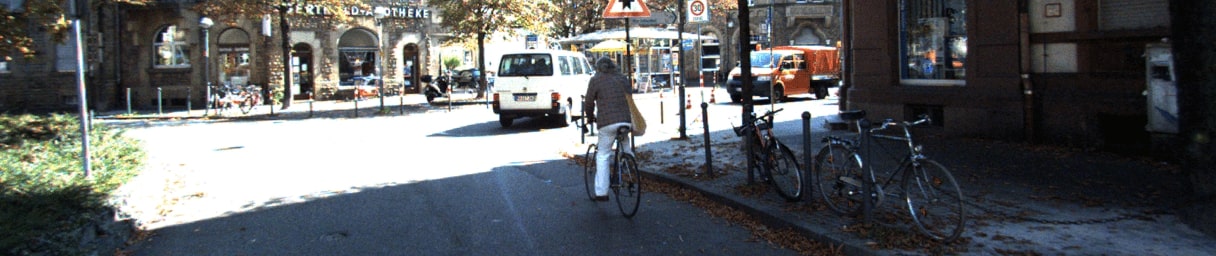}} \hspace{0.05cm}
    \subfloat{\includegraphics[width =0.38\textwidth]{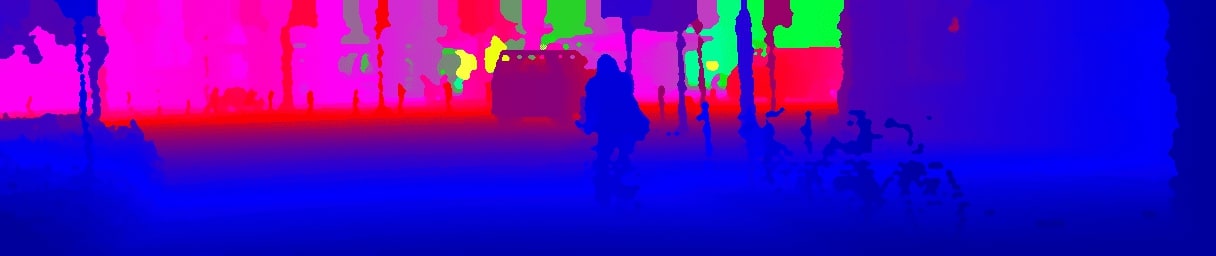}} \hfill
    \subfloat{\includegraphics[scale =0.11]{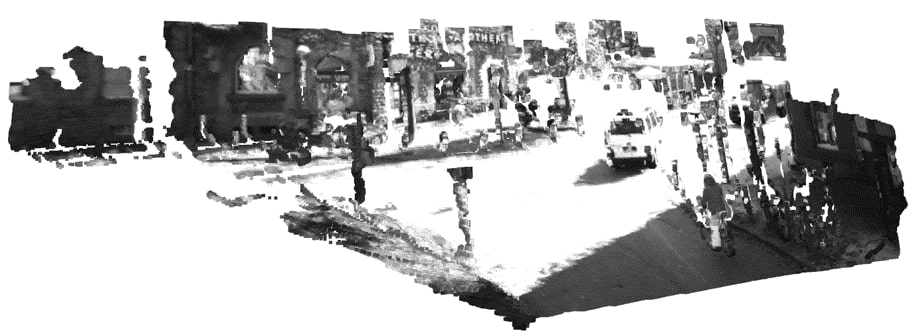}} \vspace{0.04cm}
    \subfloat{\includegraphics[width =0.38\textwidth]{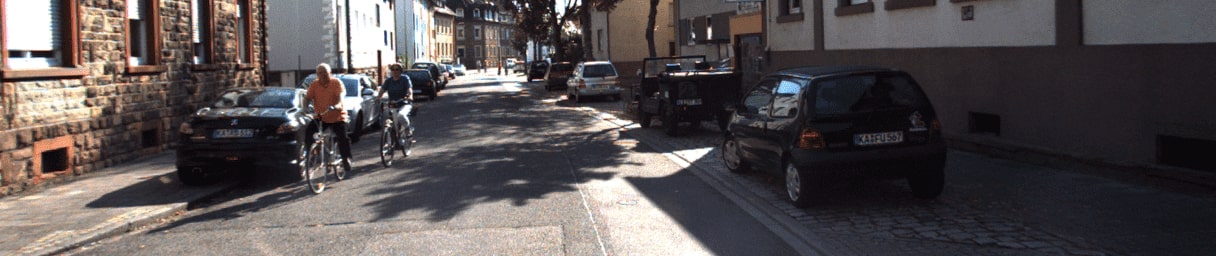}} \hspace{0.05cm}
    \subfloat{\includegraphics[width =0.38\textwidth]{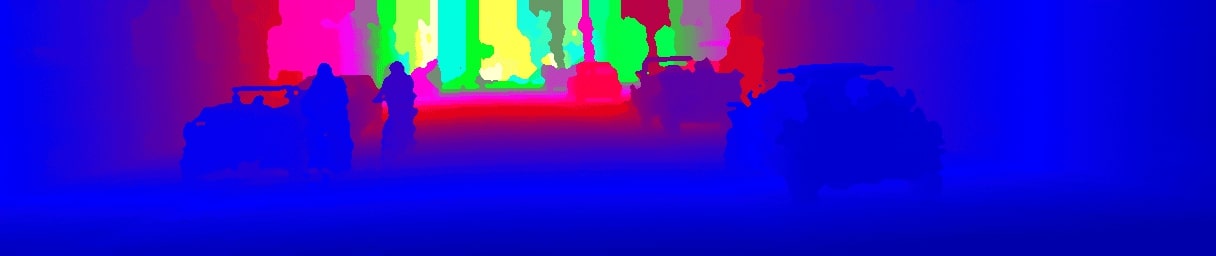}} \hfill
    \subfloat{\includegraphics[scale =0.11]{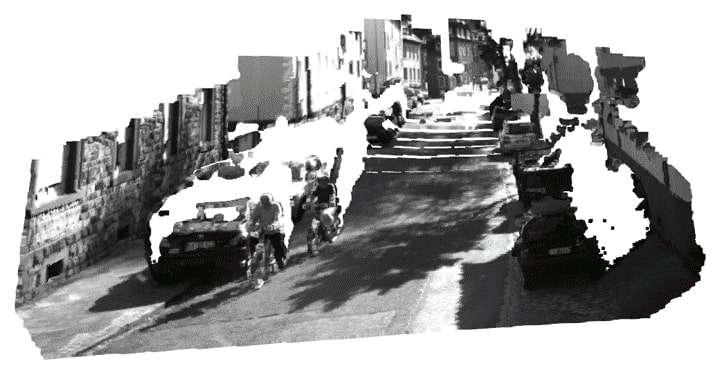}} \vspace{0.04cm}
    \subfloat{\includegraphics[width =0.38\textwidth]{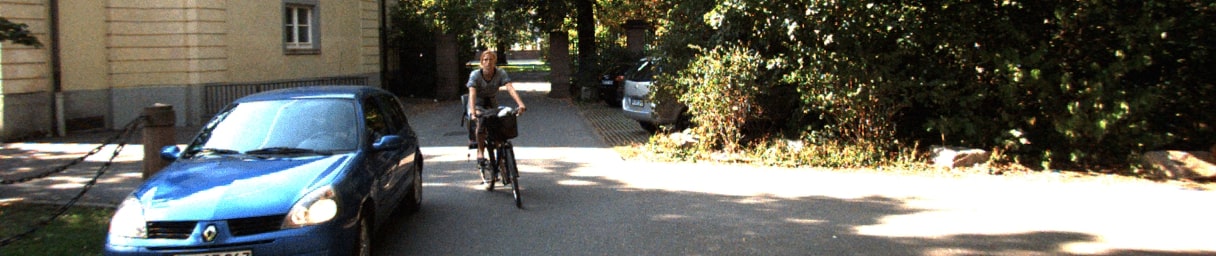}} \hspace{0.05cm}
    \subfloat{\includegraphics[width =0.38\textwidth]{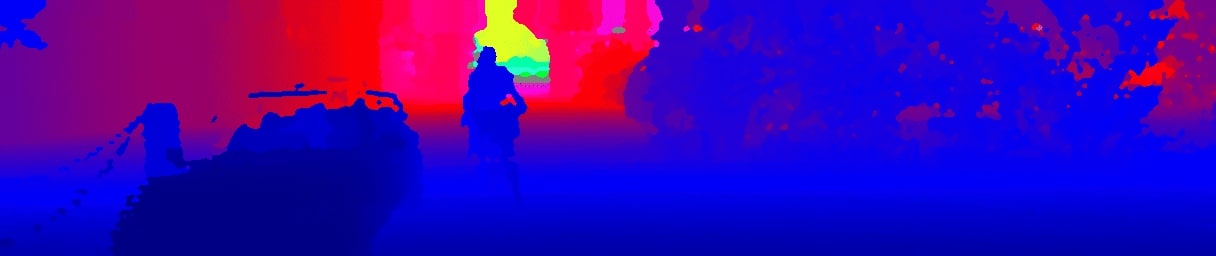}} \hfill
    \subfloat{\includegraphics[scale =0.11]{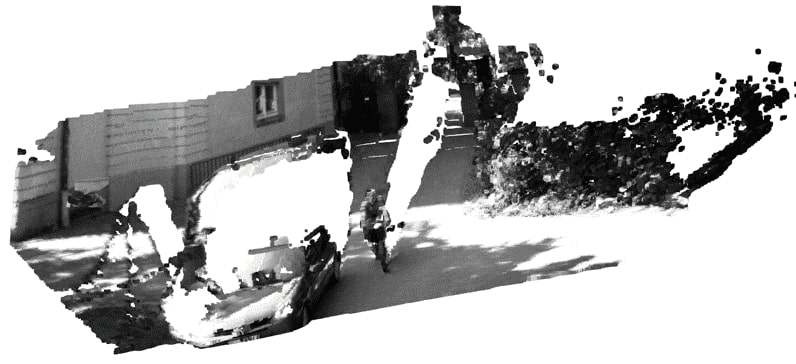}} \vspace{0.04cm}
    \subfloat{\includegraphics[width =0.38\textwidth]{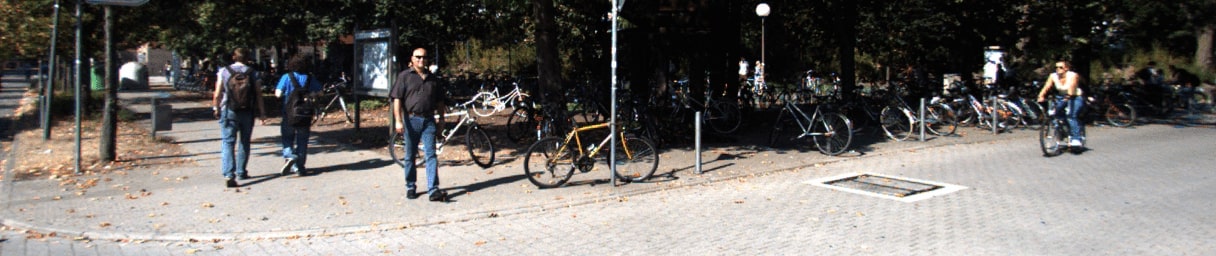}} \hspace{0.05cm}
    \subfloat{\includegraphics[width =0.38\textwidth]{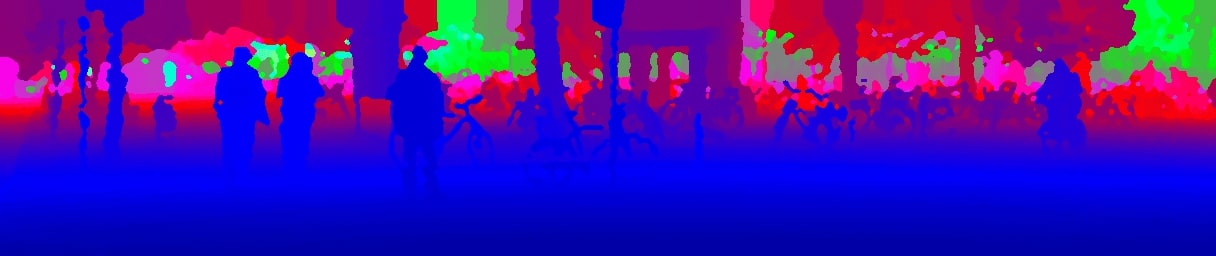}} \hfill
    \subfloat{\includegraphics[scale =0.11]{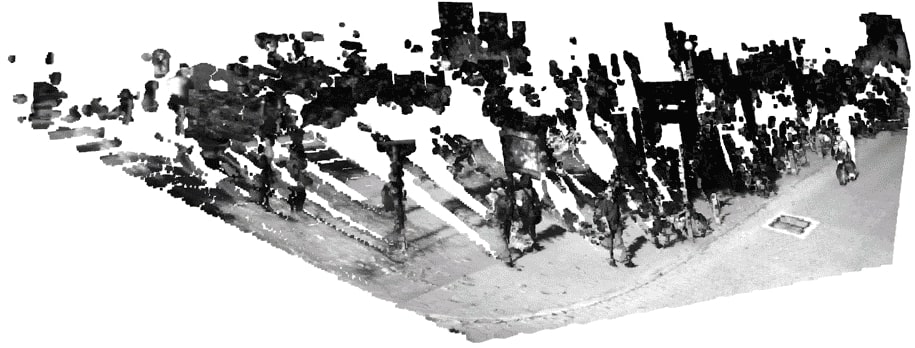}} \vspace{0.04cm}
    \subfloat{\includegraphics[width =0.38\textwidth]{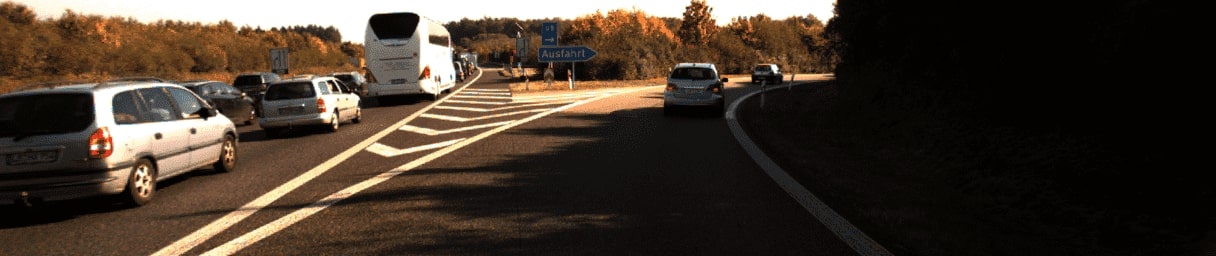}} \hspace{0.05cm}
    \subfloat{\includegraphics[width =0.38\textwidth]{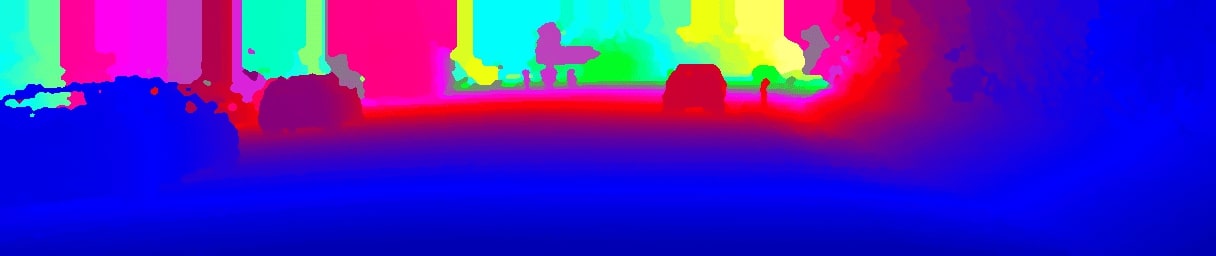}} \hfill
    \subfloat{\includegraphics[scale =0.09]{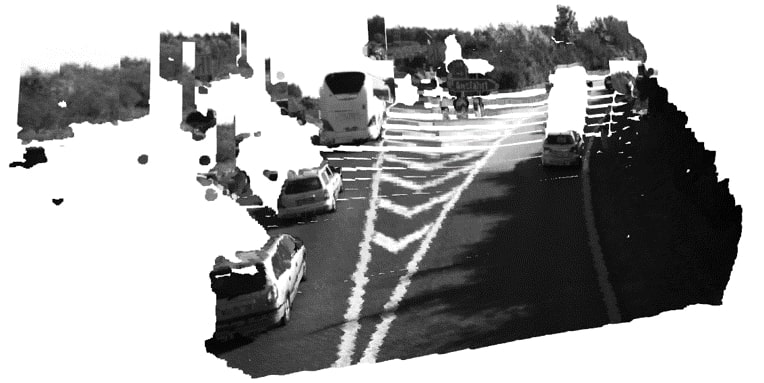}} \vspace{0.04cm}
    \subfloat{\includegraphics[width =0.38\textwidth]{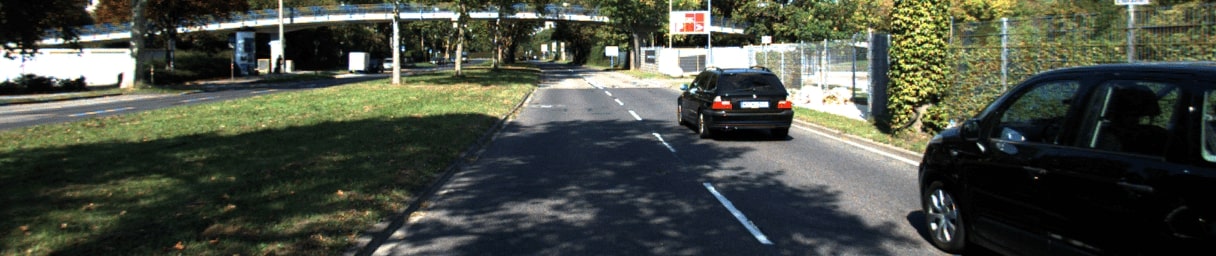}} \hspace{0.05cm}
    \subfloat{\includegraphics[width =0.38\textwidth]{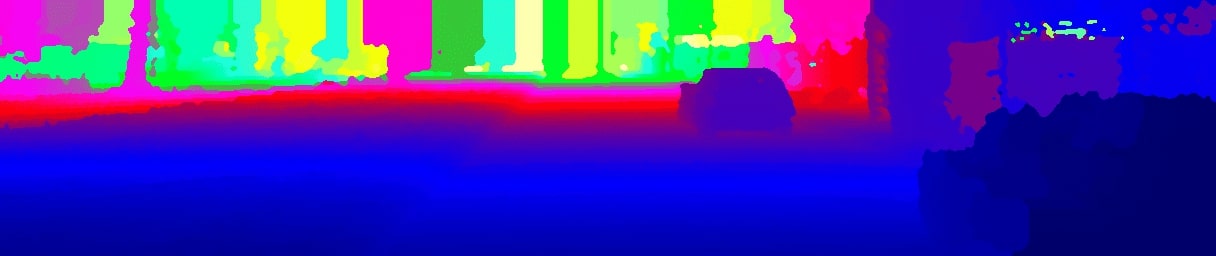}} \hfill
    \subfloat{\includegraphics[scale =0.07]{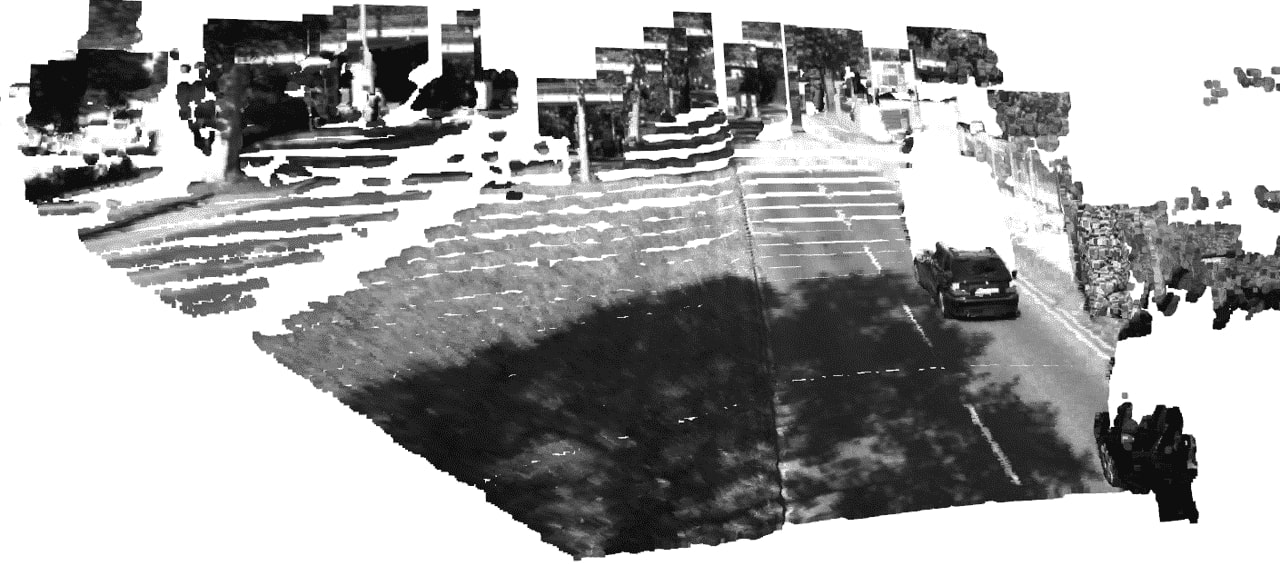}} \vspace{0.04cm}
    \subfloat{\includegraphics[width =0.38\textwidth]{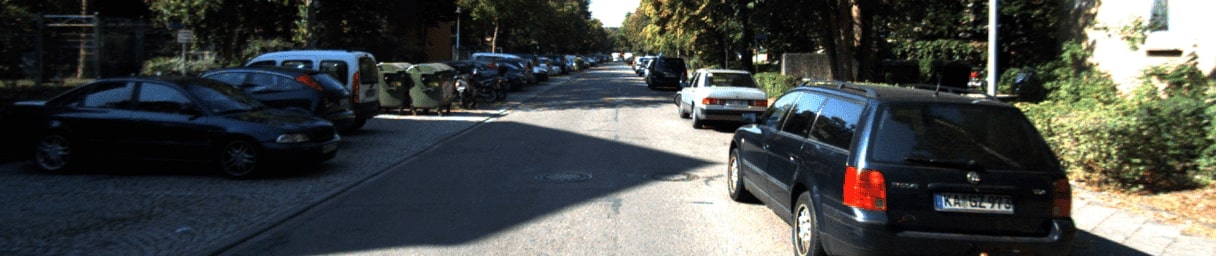}} \hspace{0.05cm}
    \subfloat{\includegraphics[width =0.38\textwidth]{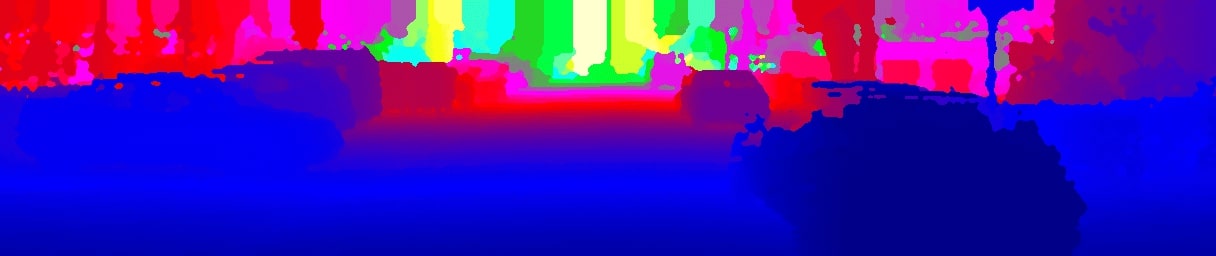}} \hfill
    \subfloat{\includegraphics[scale =0.08]{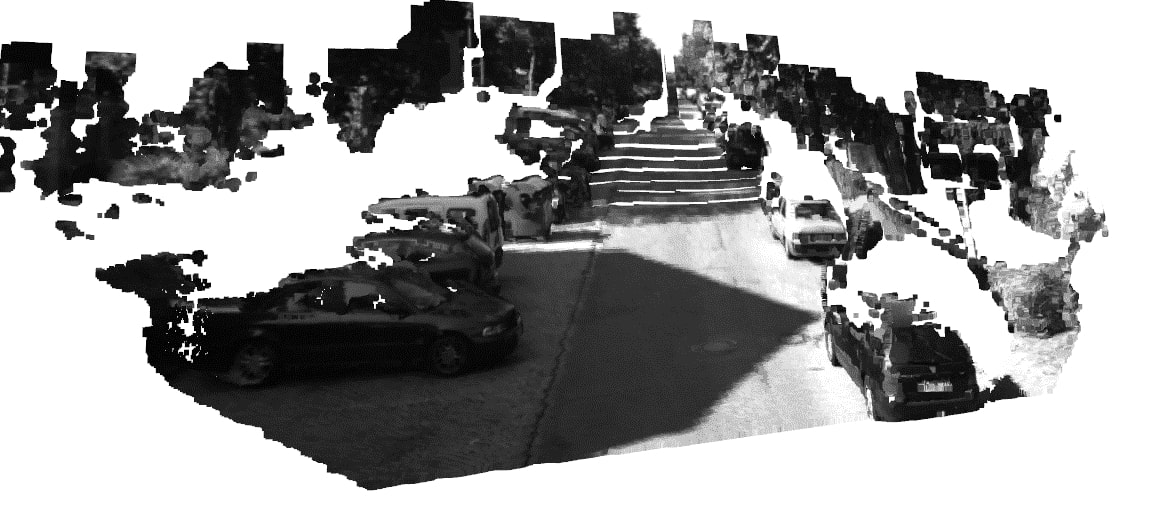}} \vspace{0.04cm}
    \subfloat{\includegraphics[width =0.38\textwidth]{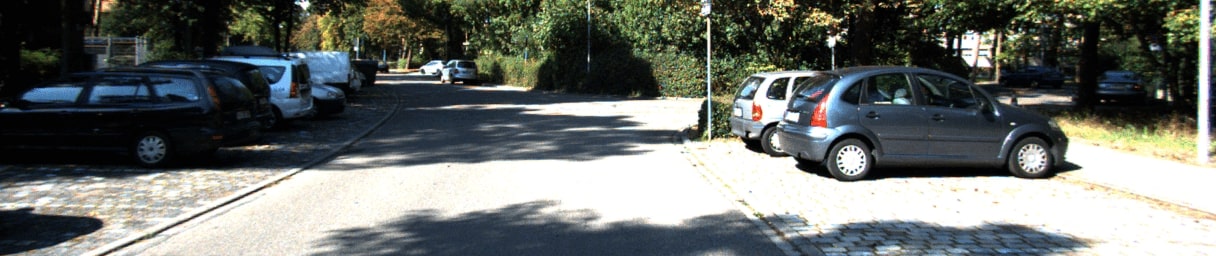}} \hspace{0.05cm}
    \subfloat{\includegraphics[width =0.38\textwidth]{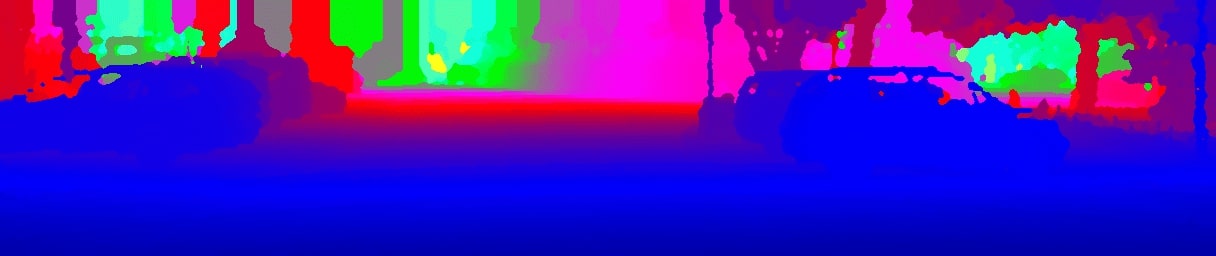}} \hfill
    \subfloat{\includegraphics[scale =0.091]{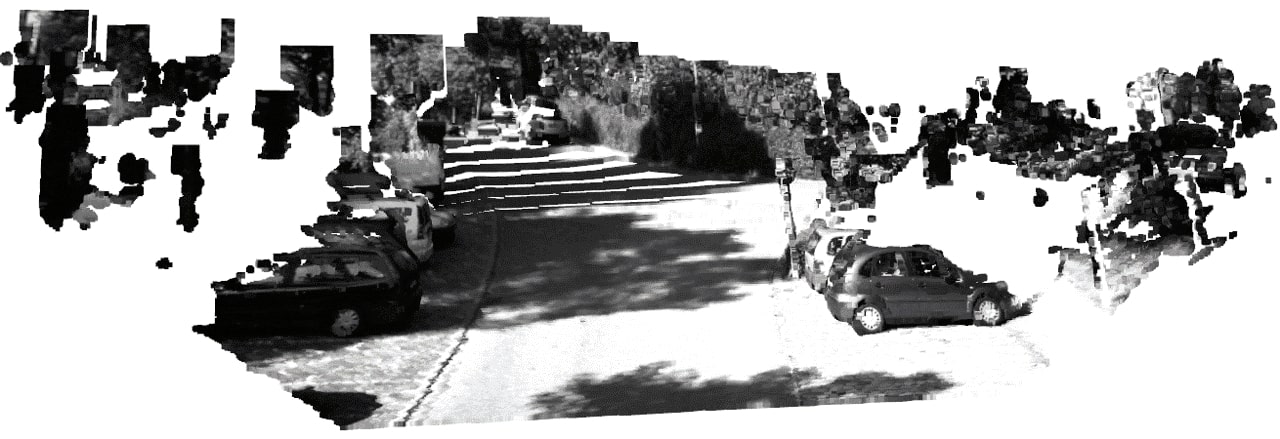}} \vspace{0.04cm}
    \addtocounter{subfigure}{-24}
    \subfloat[image]{\includegraphics[width =0.38\textwidth]{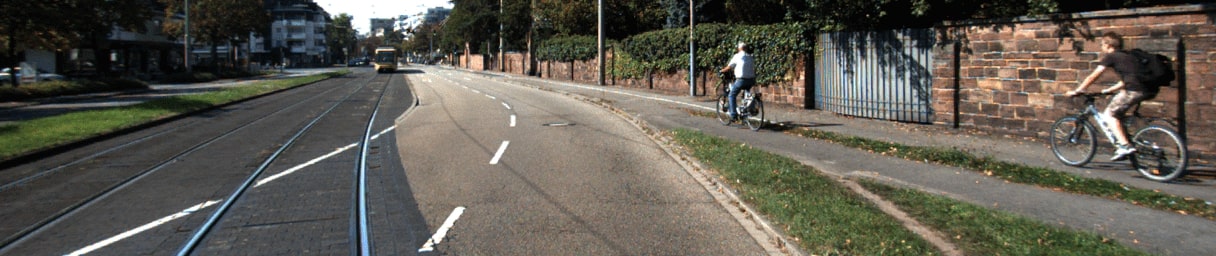}} \hspace{0.05cm}
    \subfloat[depth map]{\includegraphics[width =0.38\textwidth]{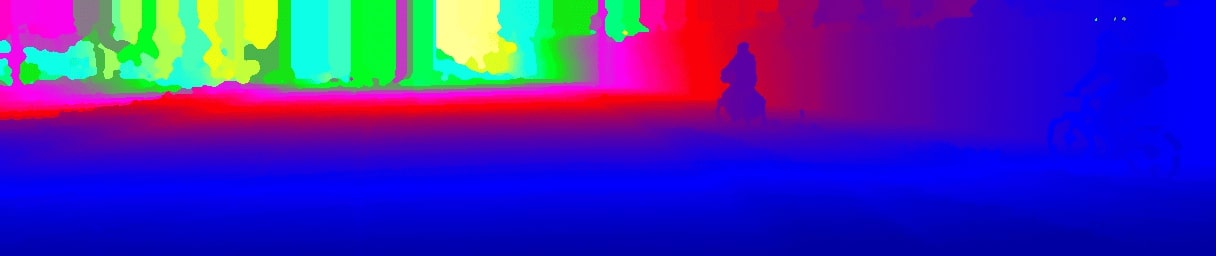}} \hfill
    \subfloat[3D view]{\includegraphics[scale =0.090]{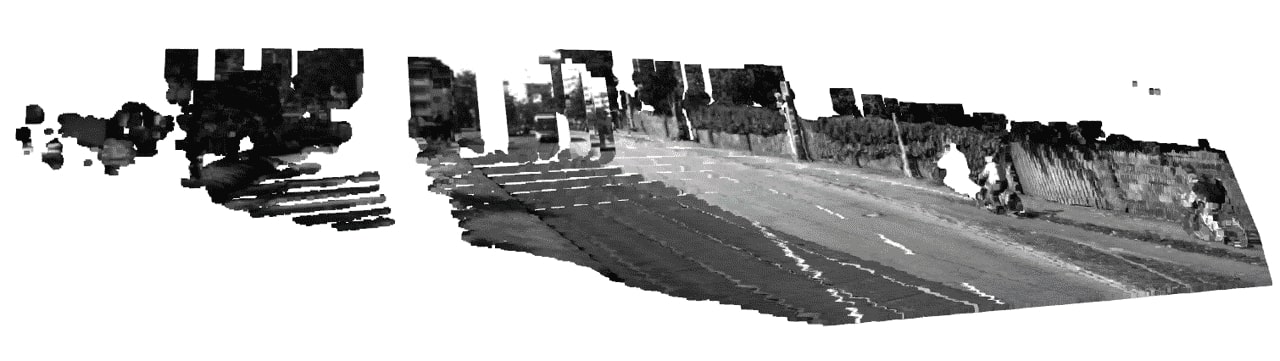}} \vspace{0.04cm}
    \caption{We display the resulting depth maps from our PDC for several examples. The row with the 3D views show that no flying pixels are generated.}
    \label{more_results}
\end{figure*}

To overcome these problems we introduce a novel geometry-based depth completion approach.
The basic assumption of our method is that depth values on an object take similar values. 
Only at the transition between objects (e.g. foreground vs. background) larger discontinuities of depth values may occur.
Therefore, one of the key tasks is to identify regions that are likely to have similar depth values.
To do this, we subdivide the corresponding RGB image into superpixels.
In order to find corresponding superpixels belonging to the same object, we use a cost map. 
Depth values can be interpolated within a cluster of associated superpixels. 
In our evaluation, we can show both the influence of the individual processing steps and the overall performance of our method.
More depth maps and 3D views of our PDC can be found in Figure \ref{more_results}.

\section{Related Works}

\begin{figure}[t]
\subfloat[RGB Input]{\includegraphics[width =0.49\columnwidth]{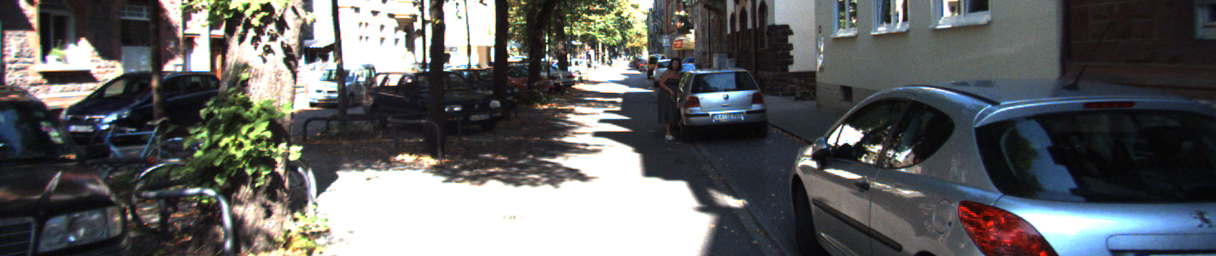}} \hfill
\subfloat[LiDAR Input] {\includegraphics[width =0.49\columnwidth]{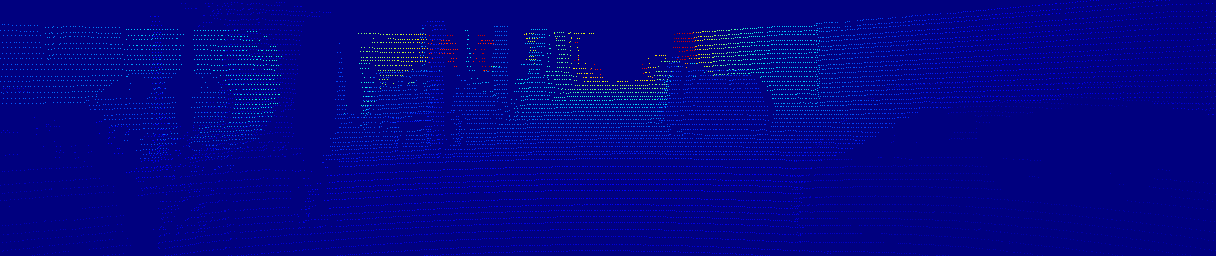}}\\
\subfloat[Revisiting \cite{cnn}] {\includegraphics[width =0.49\columnwidth]{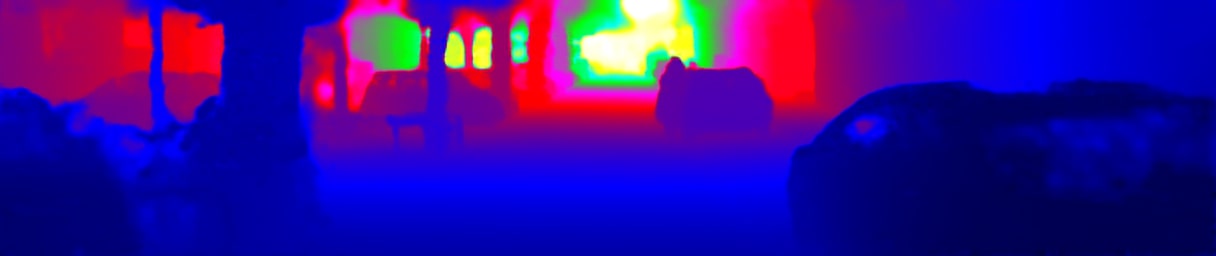}}\hfill
\subfloat[PDC (ours)] {\includegraphics[width =0.49\columnwidth]{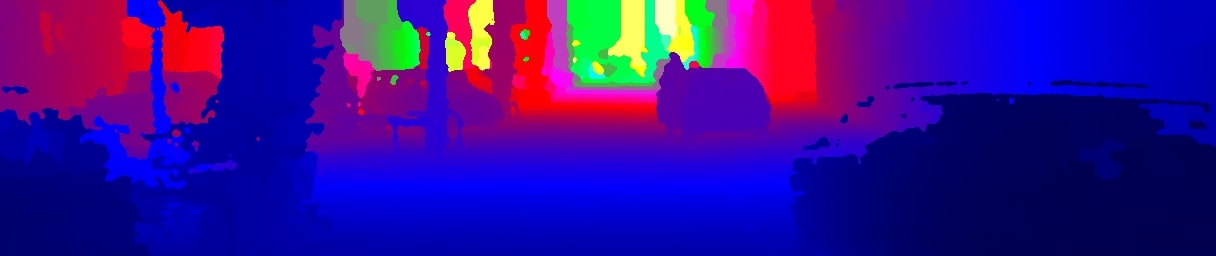}}\\
\subfloat[3D view of Revisiting \cite{cnn}] {\includegraphics[width =0.49\columnwidth]{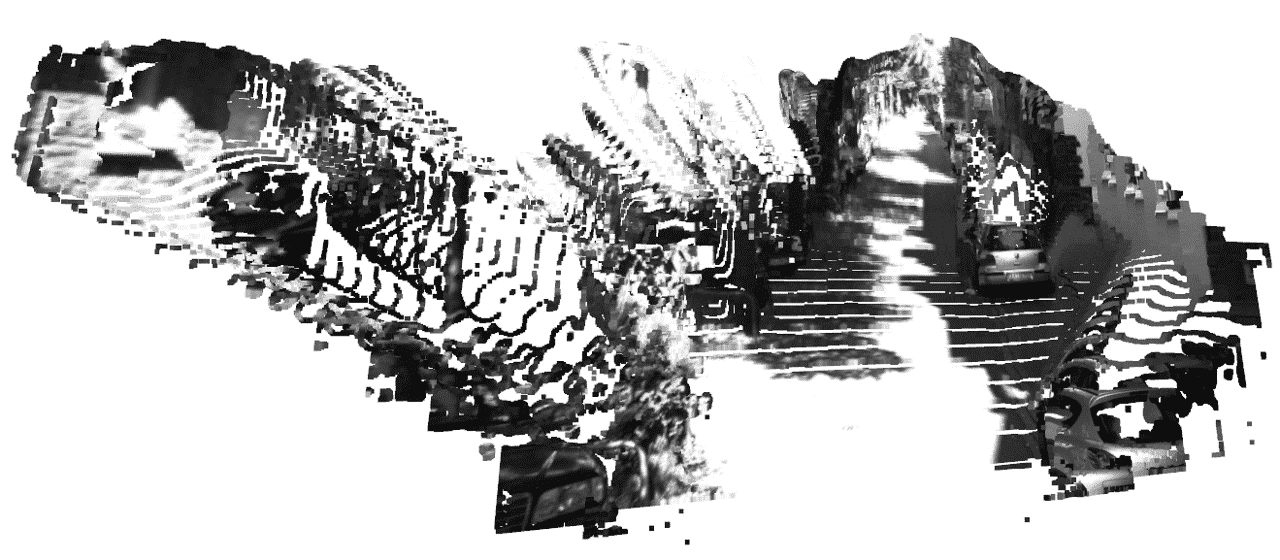}}\hfill
\subfloat[3D view of PDC (ours)] {\includegraphics[width =0.49\columnwidth]{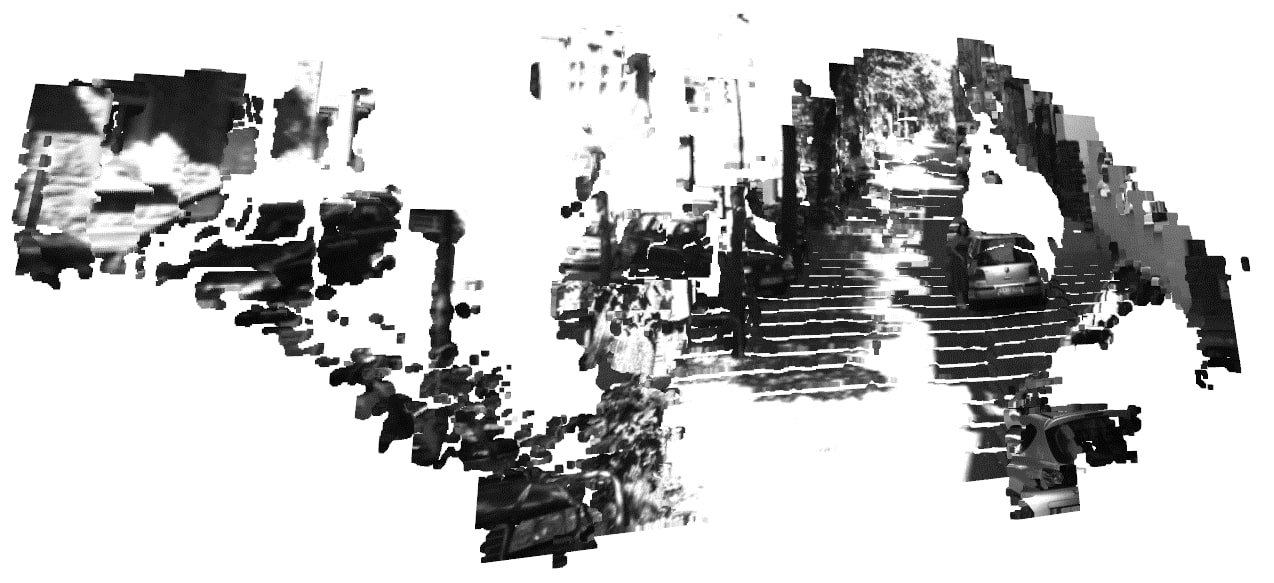}}
\caption{We propose a novel piecewise depth completion (PDC) utilizing superpixels, which -- in contrast to CNN-based methods \cite{cnn} -- effectively avoids the occurrence of flying pixels.
}
\label{teaser}
\end{figure}

\textbf{Neural networks}. 
Popular techniques for depth completion of sparse LiDAR inputs involve neural networks, which are able to learn the contours of objects and to approximate the corresponding depth values.
Particularly suitable are Convolutional Neural Networks (CNNs) such as the sparsity-invariant CNNs \cite{nncnn} in the context of depth-only completion, relying on observable depth points only.  
In the so-called image guided depth completion the sparse depth is enriched by RGB images as additional input, allowing to infer additional information which is otherwise not accessible. For instance, in the method \textit{Revisiting} \cite{cnn} the depth completion with sparsity-invariant is assisted with RGB images as additional input. 
Further methods using RGB images are \textit{Sparse to dense} \cite{sparsetodense}, \textit{DFuseNet} \cite{dfuse}, \textit{DFineNet} \cite{dfine} and \textit{Deep Lidar} \cite{deeplidar}. 
Alternatively, one can combine RGB images and LiDAR input to train the CNNs for an interpolation approach of depth values. 
For example, based on these two inputs \textit{MorphNet} \cite{morphnet} learns suitable morphological operations in order to complete the depth maps, while \textit{NConv-CNN} \cite{nconv} and \textit{SSGP} \cite{ssgp} learn the parameters of a regression model and for interpolation, respectively. 
Finally, the methods \textit{Spade-sd} \cite{sparsesd} and \textit{ADNN} \cite{adnn} both overcome sparsity as follows. The former uses a particular sparse training strategy while the latter employs techniques from compressed sensing.    

Even though the CNN-based depth completion methods produce good results in many respects, they go along with various issues. First of all, those CNNs are typically trained on specific training data sets, which may not necessarily be sufficiently representative for their eventual application. Hence, when applying these trained CNNs to new data sets from different scenes, their performance often turns out to be less robust. Furthermore, these CNNs are usually trained with the goal of minimizing only a specific metric, which may not suffice to ensure a satisfactory overall result. For instance, one often chooses to minimize the root mean square error (RMSE). However, a simultaneous reduction of the mean absolute error (MAE) is not guaranteed. Finally, we want to emphasize the qualitative problem of flying pixels occurring in regions with discontinuities, induced by the convolutional layers and the approximation of depth values. This is a common problem of CNN-based methods, resulting in a manifestly erroneous depth image (see Figure \ref{teaser} (e)). 

\textbf{Interpolation in flow estimation}. 
Many fields of computer vision face the problem of dealing with sparse data sets. In particular, the problem of flow estimation shares similarities with the challenges we face in depth completion. Specifically, there are two well-known methods for optical flow estimation, namely \textit{Epic Flow} \cite{epic} and \textit{RIC-FLow} \cite{ric}, which are based on interpolation of the matches without relying on neural networks. \textit{RIC-FLow} does not use the raw matches directly but generate a superpixel flow from input matching to improve the efficiency of the model estimation. This concept was also adapted for scene flow estimation from sparse LiDAR and RGB image input \cite{battrawy2019lidar}.
Our approach, which we describe in detail in Section \ref{Sec: Proposed Method}, applies the superpixel method in the context of depth completion.   

\textbf{Depth completion without neural networks}. To the best of our knowledge there is currently only one depth completion method on the KITTI benchmark \cite{nncnn},
which does not use neural networks. This method is called \textit{IP-Basic} \cite{ip}, which uses traditional image processing operations to produce a dense depth map. It purely relies on the LiDAR input and hence avoids the potential risk of being optimized to some specific data sets only.
In addition, \textit{IP-basic} uses the Gaussian and the bilateral blur for additional refinements. By selecting a suitable filter this method also allows to control the problem of flying pixels. However, an open issue is that \textit{IP-basic} yields an incorrect interpolation of sparse regions if there are objects in the near proximity. This problem will be discussed in more detail in \ref{Sec: Qualitative Evaluation}. 

Moreover, Buyssens \textit{et al}.~\cite{inpaint} address the problem of inpainting occlusion holes in depth maps that occur when synthesizing virtual views of a RGB-D scene. Their solution is based on using superpixels to determine missing depth values in depth maps in the context of occlusion situations. However, since our use case is the depth completion on the KITTI benchmark the results of our method are compared to \textit{IP-basic}.


\textbf{Own contributions}.
In this paper we develop a novel geometry-based depth completion method, where the depth maps are generated without the use of neural networks.
Therefore, we automatically circumvent most of the previously mentioned problems of CNN-based methods. 
Our method makes use of the image processing operations of \textit{IP-Basic} \cite{ip}.
In order to overcome the problem of wrong interpolations of sparse regions we take inspiration from \textit{RIC-Flow} \cite{ric} and introduce superpixels to geometry-based depth completion. We use superpixels in order to perform piecewise interpolation of the segments of the LiDAR input with similar depth values. 
Finally, the problem of flying pixels is solved by the choice of a suitable filter. 

\section{Proposed Method} \label{Sec: Proposed Method}

In this section, we present our approach for the geometry-based piecewise depth completion (PDC) using superpixels. 
Given an RGB image as a reference and the sparse LiDAR input, our goal is to generate a dense depth map while avoiding the notorious problem of flying pixels known from CNN-based depth-completion methods.
At the same time we want to produce results, which are qualitatively and quantitatively (based on RMSE and MAE) able to compete with the state of art methods of depth-completion. Our proposed method is summarized in Algorithm \ref{algo}.

\begin{algorithm}[H]
 \caption{Summary of our algorithm PDC}
 \label{algo}
\begin{algorithmic}[1]
\STATE Inverse LiDAR input
\STATE Dilation LiDAR input
\STATE Initialization $G(V,E)$
\STATE Initialization $cost map$
\STATE Initialization empty $mask$
\FOR{$s_i$ in $V$}
\FOR{$s_j$ with in $(s_i,s_j) \in E$}
\STATE Apply superpixel to $cost map$
\STATE Compute $\epsilon$($s_i$,$s_j$)
\IF{$\epsilon \leq \tau$}
\STATE Add $s_j$ to shortlist $L_i$
\STATE Compute cost $C(s_i,s_j)$ for $s_j \in L_i$
\ENDIF
\ENDFOR
\STATE Construct superpixel set from ranking $L_i$ with respect to cost.
\STATE Morphological close on superpixel set. 
\STATE Dilation on superpixel set.
\FOR{$s_j$ in superpixel set}
\STATE Compute median of $s_j$
\STATE Fill invalid values of $s_j$ with median
\ENDFOR
\STATE Copy superpixel set to \textit{mask}
\ENDFOR
\STATE Additional refinements
 \end{algorithmic}
\end{algorithm}

\subsection{Initialization} \label{sec: Initialization}

Our approach uses the given information of the LiDAR input as well as the image. 
In order to distinguish the depth values close to zero from invalid pixels, we invert the LiDAR input and subsequently use dilation to reduce its sparsity as proposed by Jason Ku \textit{et al}.~\cite{ip}. 
We assume that separate objects in the depth maps mostly consist of the same color but typically differ from the neighboring regions. Based on this assumption we segment the reference image into superpixels $s_i$ using the segmentation method \textit{SEED} \cite{seed}. 

\begin{figure}[t]
\includegraphics[width=\columnwidth]{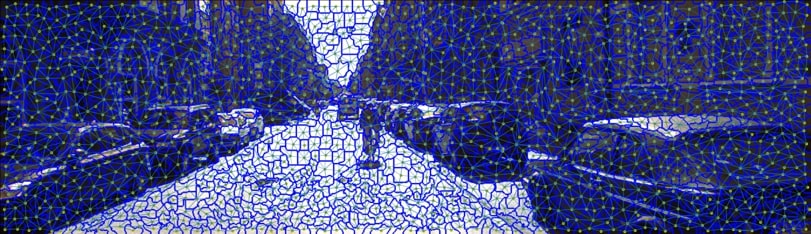}
\caption{We segment our image into superpixels, which give naturally rise to an undirected graph, where neighbors are connected by an edge. Upon using a suitable cost map it is then possible to identify the best fitting neighbors. This is the foundation for the interpolation of depth values within a set of associated superpixels.}
\label{graph}
\end{figure}

Following \cite{ric}, we introduce an undirected graph $G(V,E)$, where $V$ is the set of all superpixels $s_i$ and $E$ is the set of edges $(s_i,s_j)$ between neighboring superpixels $s_i$ and $s_j$. An example of such a graph is shown in Figure \ref{graph}.
The Euclidean distance $D(s_i,s_j)$ of two neighboring superpixels $s_i$ and $s_j$ is given by
\begin{equation}\label{distance}
D(s_i,s_j) = \sqrt{(x_i-x_j)^2+(y_i-y_j)^2} \ ,
\end{equation}
where $(x_i,y_i)$ and $(x_j,y_j)$ are the coordinates of the most centered pixels of superpixels $s_i$ and $s_j$, respectively. The distance $D(s_i,s_j)$ will be used to penalize large distant superpixels.

The detection of the best fitting neighbor requires the use of cost maps.
Y.~Hu \textit{et al.} \cite{ric} use the result of the structured edge detector SED \cite{sed} as their cost map. Alternatively, a cost map can be obtained from the reference image in gray scale. 
Since we already use the color information to segment the reference image it seems natural to also use the gray scale for the detection of the best fitting neighbors. Tests have shown that they produce similar results but the time consumption of the initialization of SED is significantly higher than the transformation to gray scale. These results can be found in Table \ref{t_time} of Section \ref{Sec: Evaluation}. 
Consequently, we henceforth use the gray scale image to obtain the cost map.

\begin{figure}[t]
\subfloat[Inlier]{\includegraphics[width =0.475\columnwidth]{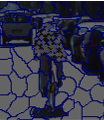}} \hfill
\subfloat[Morphological close] {\includegraphics[width =0.475\columnwidth]{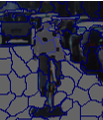}}\\
\subfloat[Dilation] {\includegraphics[width =0.475\columnwidth]{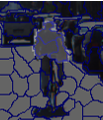}}\hfill
\subfloat[Cut] {\includegraphics[width =0.475\columnwidth]{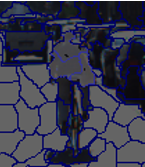}}\\
\centering
\subfloat[filling with median] {\includegraphics[width =0.475\columnwidth]{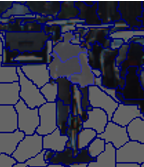}}
\caption{Our proposed method identifies the LiDAR inputs which belong to the same object (a). It uses a morphological close function to fill in the gaps (b), followed up by dilation in order to close small gaps (c). In the last step we cut out the set (d) and fill the remaining empty pixels with the median of the corresponding segment (e).  }
\label{inlier}
\end{figure}

\subsection{Identification of the best fitting neighbors}\label{sec: Neighbors}

A superpixel $s_i$ can be interpreted as a cluster of many pixels. Each of those pixels obtains gray scale values, denoted by $v^{(i)}_k$, from the image. Given two neighboring superpixels $s_i$ and $s_j$ we compute the MAE (henceforce denoted by $\epsilon$) for a sample of pixels as follows: 
\begin{equation}
\epsilon(s_i,s_j) = \frac{1}{n}\sum_{k=1}^{n}\lvert v^{(i)}_k-v^{(j)}_k \rvert
\end{equation}
Here, $n$ is the sample size defined by $n = \min(|s_i|,|s_j|)$, where $|s_i|$ denotes the number of pixels that make up superpixel $s_i$. 

Based on $\epsilon(s_i,s_j)$ we construct a shortlist $L_i$ of best fitting neighbors for $s_i$ as follows: Any neighboring superpixel $s_j$ is added to $L_i$ if and only if $\epsilon(s_i,s_j) \leq \tau$, where $\tau>0$ is some threshold.   
Now we rank these shortlisted superpixels according to some cost $C$, which disfavors large distant superpixels. Following \cite{ric} we define the cost $C$ for $s_j$ with respect to $s_i$ by
\begin{equation}
C(s_i,s_j) = W \cdot \epsilon(s_i,s_j) \qquad \forall ~ s_j \in L_i \ ,
\end{equation}
where the weight function $W$ is given by $W = \exp(-D(s_i,s_j)/\alpha)$ with some regularization parameter $\alpha>0$. 
Therefore, each of the superpixels $s_j \in L_i$ has a corresponding cost $C(s_i,s_j)$. The two superpixels of $L_i$ with the smallest cost are henceforth called the best fitting neighbors.   
The superpixels and their best fitting neighbors will be called superpixel set for the remainder of this paper.

\subsection{Interpolation with morphological operations}

Having constructed the superpixel sets we are able to cluster the pixels which possess similar depth values as seen in Figure \ref{inlier} a). Therefore, we are in the position to make use of morphological operations for our interpolation approach. 

We use the morphological close function to merge the pixels that are close to each other. However, large clusters of empty pixels are not filled by this step. To fill them, we use the dilated values of the superpixel set.
Inserting the dilated values causes the pixels to go beyond the boundaries of our superpixel set as shown in Figure \ref{inlier} (c). To fix this issue, the superpixels are cut out one by one along the boundaries. By doing this, we remove the overhanging pixels as illustrated in Figure \ref{inlier} (d). At the same time we remove the depth values that were taken from the dilation from a neighboring superpixel. This ensures that if a neighbor was incorrectly declared an inlier, no incorrect depth data is transferred to other superpixels.
Inspired by RIC-flow \cite{ric}, the remaining empty pixels of this process are filled with the median of their respective superpixel as shown in Figure \ref{inlier} (e).
This iterative process is done for each superpixel set.

\subsection{Additional refinements}\label{refine}

Our approach only uses the existing depth values from the LiDAR input. 
However, larger objects may not be fully captured by the sensor, i.e.~these objects are not fully covered by the depth map. 
To complete the depth map, the top row pixels are extrapolated to the top point of the image plane as is done in IP-Basic \cite{ip}. This step is optional and changes the result of the evaluation only insignificantly.

Furthermore, there can still be invalid pixels in the area covered by the sensor. This is a consequence of particularly sparse regions in the original LiDAR input. In order to fill the remaining invalid pixels, we follow the method of IP-basic \cite{ip} and overwrite the invalid pixels using dilated pixel values. 

So far, our approach does not generate any flying pixels. However, this is no longer guaranteed once we further optimize the depth maps by including filters. The two relevant filters for this paper are the Gaussian blur and the bilateral blur. The former filter can be used to reduce the noise in our depth maps, but it does not have edge preserving properties. Hence, this filter will produce flying pixels in regions with an edge. In contrast, the bilateral blur does preserve edges. Therefore, our PDC method can be combined with a bilateral blur filter to further optimize the depth maps without running into the notorious problem of generating flying pixels.

\section{Evaluation} \label{Sec: Evaluation}

In this section, we show the design decisions of our PDC algorithm in a detailed ablation study. 
In addition, we perform a quantitative as well as qualitative evaluation on the KITTI dataset \cite{nncnn}.

\subsection{Ablation Study}

One part of our contribution is the search of the best possible neighbors for our interpolation (see Section \ref{sec: Neighbors}). 
In order to show that this step has a significant positive impact on the evaluation results, we present in Table \ref{t_ablation} the results of our method with and without the use of a superpixel set where each superpixel is interpolated separately. 
We also show the results of the two filters as well as with and without extrapolation (see Section \ref{refine}) of the pixels to the top of the frame. 
This evaluation shows that the interpolation with a superpixel set generates consistently better results with regard to RMSE and MAE. 

\begin{table}[t]
\caption{We compare our PDC with different configurations on the public KITTI training data \cite{nncnn}. 
It shows that the formation of superpixel sets positively affects the results.
}
\centering
\adjustbox{max width=\columnwidth}{%
\begin{tabular}{@{}cccc|ll@{}}
\toprule
\multicolumn{1}{l}{\begin{tabular}[c]{@{}l@{}} bilateral \\ blur\end{tabular}} & \multicolumn{1}{l}{\begin{tabular}[c]{@{}l@{}}Gaussian\\blur\end{tabular}} & \multicolumn{1}{l}{extrapolated} & \multicolumn{1}{l|}{superpixel set} & RMSE{[}mm{]} & MAE{[}mm{]} \\ \midrule
\xmark                                 & \xmark                                & \xmark                                     & \cmark & 1426.163 & 313.368             \\
\xmark                                 & \xmark                                & \cmark                                    & \xmark                                                           &     1429.664         &      294.238       \\
\xmark                                 & \xmark                                & \cmark                                     & \cmark                                                           &   1420.567           &    312.829         \\
\xmark                                 & \cmark                                & \xmark                                     & \xmark                                                           &   1310.346           &    297.007         \\
\xmark                                 & \cmark                                & \xmark                                     & \cmark                                                           &  1307.856             &     295.152        \\
\xmark                                 & \cmark                                & \cmark                                     & \xmark                                                           &    1303.344          &       296.389      \\
\xmark                                 & \cmark                                & \cmark                                     & \cmark                                                           &    1286.555          &       293.221      \\
\cmark                                & \xmark                                & \xmark                                     & \xmark                                                           &    1437.159          &      294.863       \\
\cmark                                & \xmark                                & \xmark                                     & \cmark                                                           &     1434.791         &      293.312       \\
\cmark                                & \xmark                                & \cmark                                     & \xmark                                                           &     1429.664         &      294.238       \\
\cmark                                & \xmark                                & \cmark                                     & \cmark                                                           &    1412.831          &    291.449         \\ \bottomrule
\end{tabular}}
\label{t_ablation}
\end{table}

We can observe that the Gaussian blur achieves the best RMSE, but the bilateral blur achieves the best MAE. 
The reason for this is that the RMSE accounts for larger errors in the depth map more than the MAE.
The conclusion is that the Gaussian blur is able to reduce the noise of larger flat regions more efficiently than the bilateral blur, while it gives a worse result in edge regions that can be considered as small errors. 
From this evaluation, we can conclude that the bilateral blur is the more suitable filter because it preserves more precise contours of objects. 

In addition, we compare the time required to initialize the cost maps (see section \ref{sec: Initialization}) for a depth map in Table \ref{t_time}. 
It can be seen that although the error metrics are almost the same, the time required by SED \cite{sed} is significantly higher.

\begin{table}[t]
\caption{We compare the results of the two cost maps. 
It shows that the gray scale image achieves similar results to SED but initializes much faster.
For this result the following environment was used: 1 core @3.6GHz (Python).}
\adjustbox{max width=\columnwidth}{%
\begin{tabular}{@{}lccc@{}}
\toprule
                                      & RMSE {[}mm{]} & MAE {[}mm{]} & extra time per image {[}s{]} \\ \midrule
\multicolumn{1}{l|}{SED \cite{sed}}              & 1286.536      & 293.259      & 1.38688                      \\
\multicolumn{1}{l|}{gray scale image} & 1286.555      & 293.221      & 0.0109                      
\end{tabular}}
\label{t_time}
\end{table}
It is important to note that we have not yet focused our PDC on optimizing the computation time but on the quality of the depth maps. Currently, the computational step of finding each superpixel and its neighbors is computationally very intensive. The optimization of the computational time is left for future work. 

\subsection{Quantitative Evaluation}

The best method so far on the KITTI dataset, which also does not use deep learning, is the IP-Basic \cite{ip} approach. 
In Table \ref{t_ip}, we compare our new PDC against this approach.
It turns out that our method, with the additional use of an RGB image instead of just using the raw LiDAR data as input, consistently yields better results. 
The error metrics show that the Gaussian blur produces the better RMSE while the bilateral blur produces the better MAE.
\begin{table}[t]
\caption{ We compare IP-Basic \cite{ip} with PDC (ours) on the public KITTI training data \cite{nncnn}.
It can be seen that PDC (ours) with the filters used by IP-Basic, produces better results overall.
}
\centering
\adjustbox{max width=\columnwidth}{%
\begin{tabular}{@{}l|llcc@{}}
\toprule
\multicolumn{1}{c|}{} & \multicolumn{2}{c}{RMSE{[}mm{]}}              & \multicolumn{2}{c}{MAE{[}mm{]}}                                        \\
                      & Gaussian blur & Bilateral blur                & \multicolumn{1}{l}{Gaussian blur} & \multicolumn{1}{l}{Bilateral blur} \\ \midrule
IP-Basic \cite{ip}             & 1350.927      & \multicolumn{1}{l|}{1454.754} & 305.352                           & 303.699                            \\
PDC (ours)            & 1286.555      & \multicolumn{1}{l|}{1412.831} & 293.221                           & 291.449                            \\ \bottomrule
\end{tabular}}
\label{t_ip}
\end{table}

For the purpose of comparing our method quantitatively with the state of art, we submitted our results on the KITTI benchmark test \cite{nncnn}. 
We present the evaluation result in Table \ref{t_cnn} which we structure as follows. 
The CNN column describes if the pictured method uses a convolution neural network, the RGB column depicts if an RGB image is additionally used for guided depth completion and learning interpolation is used to describe if a method learns parameters to interpolate the depth values.

This evaluation shows that our method achieves similar results on the public and anonymous KITTI datasets \cite{nncnn}.
Therefore we do not have the problem of optimization and our approach is robust with different inputs.
While the best CNN based methods yield better results in respect to the RMSE, our proposed method is en par regarding the MAE. 
Deep learning optimizes the RMSE and neglects the MAE.
Therefore, the MAE is proportionally better with ours.
The same is true for IP-Basic \cite{ip}. Consequently, this observation supports our investigation into geometry-based approach to depth completion as a competitor to established deep-learning methods.

\begin{table}[t]
\caption{We compare our method with the state of the art from the KITTI benchmark \cite{nncnn}.
It shows that PDC (ours) does not perform as well as CNN-based methods in terms of RMSE, but can compete with them in terms of MAE.
}
\adjustbox{max width=\columnwidth}{%
\begin{tabular}{@{}lccccc@{}}
\toprule
  & \multicolumn{1}{l}{CNN} & \multicolumn{1}{l}{RGB} & \multicolumn{1}{l}{\begin{tabular}[c]{@{}l@{}}learning \\ interpolation\end{tabular}} & \multicolumn{1}{l}{RMSE{[}mm{]}} & \multicolumn{1}{l}{MAE{[}mm{]}} \\ \midrule
\multicolumn{1}{l|}{Deep Lidar \cite{deeplidar}}          &  \cmark   &  \cmark   &          \xmark                                                          & 758.38       & 226.50      \\
\multicolumn{1}{l|}{Revisiting \cite{cnn}}                &   \cmark  &  \cmark   &         \xmark                                                           & 792.80       & 225.81      \\
\multicolumn{1}{l|}{Sparse to Dense \cite{sparsetodense}} &   \cmark  &  \cmark   &      \xmark                                                              & 814.73       & 249.95      \\
\multicolumn{1}{l|}{SSGP \cite{ssgp}}           &   \cmark  &  \cmark   &                \cmark                                                    & 838.22      & 244.7      \\
\multicolumn{1}{l|}{DFineNet \cite{dfine}}                &   \cmark  &  \cmark   &           \xmark                                                         & 943.89       & 304.17      \\
\multicolumn{1}{l|}{Spade-sD \cite{sparsesd}}             &  \cmark   &  \cmark(\xmark)   &                     \xmark                                               & 1035.29      & 248.32      \\
\multicolumn{1}{l|}{MorphNet \cite{morphnet}}             &   \cmark  &   \cmark  &                \cmark                                                   & 1045.45      & 310.49      \\
\multicolumn{1}{l|}{DFuseNet \cite{dfuse}}                &   \cmark  &   \cmark  &                   \xmark                                                & 1206.66      & 429.93      \\
\multicolumn{1}{l|}{PDC(ours)}                                                &  \xmark   &  \cmark   &                        \xmark & 1227.96      & 288.55      \\
\multicolumn{1}{l|}{NConv-CNN \cite{nconv}}               &   \cmark  &   \cmark  &          \xmark                                                         & 1268.22      & 360.28      \\
\multicolumn{1}{l|}{IP-Basic \cite{ip}}                   &   \xmark  &   \xmark  &         \xmark                                                          & 1288.46      & 302.60      \\
\multicolumn{1}{l|}{ADNN \cite{adnn}}                     &   \cmark  &   \xmark  &          \xmark                                                         & 1325.37      & 439.48      \\
\multicolumn{1}{l|}{NN+CNN \cite{nncnn}}                  &  \cmark   &   \xmark  &        \xmark                                                           & 1419.75      & 416.14      \\
\multicolumn{1}{l|}{SparseConvs \cite{nncnn}}       &   \cmark  &   \xmark  &      \xmark                                                             & 1601.33      & 481.27     
\end{tabular}}
\label{t_cnn}
\end{table}

\subsection{Qualitative Evaluation} \label{Sec: Qualitative Evaluation}

\begin{figure}[t]
\subfloat[RGB]{\includegraphics[width =0.49\columnwidth]{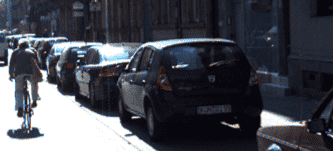}} \hfill
\subfloat[Ground truth] {\includegraphics[width =0.49\columnwidth]{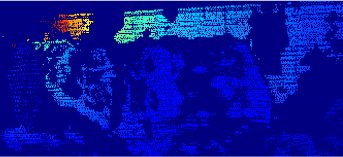}}\\
\subfloat[IP-Basic \cite{ip}] {\includegraphics[width =0.49\columnwidth]{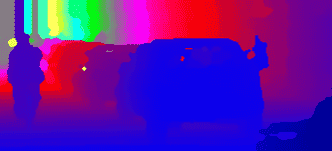}}\hfill
\subfloat[PDC (ours)] {\includegraphics[width =0.49\columnwidth]{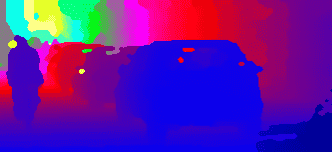}}
\caption{This figure shows that PDC (d) is able to produce a reliable result even in sparse regions, while IP-Basic (c) does not. The ground truth (b) shows that the wrong interpolation do not get evaluated.}
\label{q_ipvspdc}
\end{figure}

\textbf{IP-Basic vs. PDC (ours)}. 
In order to show that our approach handles the interpolation of sparse regions better than IP-Basic \cite{ip} we visually compare it with PDC in Figure \ref{q_ipvspdc}. 
It shows that in areas close to invalid pixels IP-Basic struggles to keep the contours of the  objects. 

These problems arise because IP-Basic does not differentiate between objects. 
Furthermore, it uses the pixels from the car to fill the gaps in the wall close to it. 
Our PDC does not have these problems because we use a piecewise interpolation approach where the used data is usually from the same object. 
Since there is no ground truth at these regions it does not impact the metric based evaluation even though a significant error in the depth map is visible. 
Thus, we were also able to show visually that our method produces significantly higher quality depth information than IP-Basic.

\begin{figure}[t]
\subfloat[RGB input]{\includegraphics[width =0.25\columnwidth]{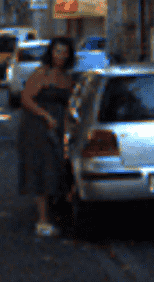}} \hfill 
\subfloat[Revisiting \cite{cnn}] {\includegraphics[width =0.25\columnwidth]{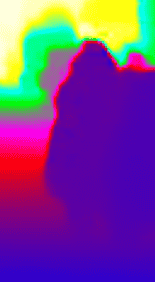}}\hfill
\subfloat[PDC (ours)] {\includegraphics[width =0.25\columnwidth]{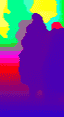}}\\
\subfloat[LiDAR input] {\includegraphics[width =0.25\columnwidth]{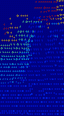}}\hfill
\subfloat[Error map of (b)] {\includegraphics[width =0.25\columnwidth]{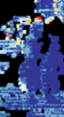}}\hfill
\subfloat[Error map of (c)]{\includegraphics[width =0.25\columnwidth]{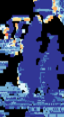}} 
\caption{This figure shows the resulting depth maps (c) and (d) from the LiDAR input (b). In the corresponding error maps (e) and (f) the blue pixels are the correct estimated depth values.}
\label{q_passant}
\end{figure}

\textbf{CNN based vs. PDC (ours)}. 
The CNN-based method Revisiting \cite{cnn} uses SI-Convolution and an RGB image as additional input. The method produces good results in terms of RMSE and MAE.  Since it can compete with the best methods on the KITTI benchmark \cite{nncnn}, we use it as the representative CNN-based method. So in order to further support our claim that our method is able to compete with CNN-based methods, we compare the resulting depth maps of Revisiting \cite{cnn} with ours.
For this purpose, we decided to select a representative image with a potential danger for a pedestrian and consider the depth map of this said person in Figure \ref{q_passant}.

It shows that PDC is able to preserve the edges of the person and can be recognized in the depth map as such. 
In the depth map of Revisiting it is harder to recognize the person since she melts together with the car which makes it harder to differentiate. 
The error maps in Figure \ref{q_passant} (e) and Figure \ref{q_passant} (f) makes this more apparent and show how the approximation of the depth values do not reflect the reality. 
Since our approach only uses the present data and does not approximate we have more depth values in the error map which correspond to the ground truth. 
We have noticed that our approach is not able to compete with CNN based methods in regions with exceptionally sparse LiDAR input but our method is a strong contender in regions with dense input. 
Since most of the dense areas are right in front of the car in a close proximity  it is obvious that these are very important regions. 
This comparison was done with all listed CNN based methods from Table \ref{t_cnn} with similar results.

\begin{figure}[t]
\centering
\subfloat[Revisiting \cite{cnn}] {\includegraphics[width =0.25\columnwidth]{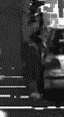}}\hspace{0.1cm}
\subfloat[PDC (ours)]{\includegraphics[width =0.25\columnwidth]{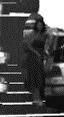}} \hspace{0.1cm}
\caption{3D view of the depth map from Figure \ref{q_passant} which shows how the flying pixels distort the pedestrian in (a) in comparison with PDC (ours) (b) where no distortion takes place}
\label{q_fly}
\end{figure}

The other issue is the flying pixels which are often generated by CNN based methods. 
The depth map from Revisiting (see Figure \ref{q_passant} (b)) shows that there are red pixels around the pedestrian. 
That means that they are farther away from the person and do not represent the reality. 
This corresponds to a transition of pixels, which are then visible as flying pixels in the 3D view.
Usually they are not considered in the evaluation because there are often no ground truths in this edge regions. 
This can be observed in the error maps in Figure \ref{q_passant} (e) and Figure \ref{q_passant} (f) where the black spots represent the absence of the ground truth values. 
In Figure \ref{q_fly} these flying pixels are visualized for this specific example and show that the pedestrian in Figure \ref{q_fly} (a) is completely distorted while it is not true for the results of our method in  Figure \ref{q_fly} (b). 
The whole depth map and the 3D view from which this detail is taken is shown in the Figure \ref{teaser}.

\section{Conclusion}
In this paper, we have introduced a new geometry-based approach for depth completion by using superpixels for a piecewise interpolation of the depth map.
We show that we can produce reliable depth maps with the segmentation of an additional RGB image into superpixels. 
The identification of suitable interpolation partners sharing similar depth values eliminates the problem of erroneous interpolation in sparse regions. 
We have also shown that our PDC avoids flying pixels as well as can compete with and outperform CNN-based depth completion methods in some regions.

\section*{Acknowledgment}
This work was funded by the Karl V\"olker Foundation in the project "KI-Fusion".
We thank Laurenz Reichardt for the fruitful discussions and his positive inputs.

\bibliographystyle{IEEEtran}
\bibliography{IEEEabrv,literatur}

\end{document}